\documentclass[runningheads]{llncs}

\usepackage[utf8]{inputenc}

\usepackage[LAE, T1]{fontenc}
\usepackage[arabic, english]{babel}
\usepackage[free]{ARfonts}

\usepackage{graphicx}
\usepackage{amsmath}
\usepackage{xcolor}

\begin{document}
\title{An open access NLP dataset for Arabic dialects : data collection, labeling, and model construction}
%
%
\author{ElMehdi Boujou \and
Hamza Chataoui \and
 Abdellah El Mekki \and
Saad Benjelloun \and
Ikram Chairi \and
Ismail Berrada}
\authorrunning{Boujou et al.}
\titlerunning{An open access NLP dataset for Arabic dialects}
%
\institute{Mohamed VI Polytechnic University (UM6P), Lot 660, Hay Moulay Rachid, Ben Guerir 43150, Morocco.\\
}

\maketitle              
\begin{abstract}
Natural Language Processing (NLP) is today a very active field of research and innovation. Many applications need however big sets of data for supervised learning, suitably labelled for the training purpose. This includes applications for the Arabic language and its national dialects. However, such open access labeled data sets in Arabic and its dialects are lacking in the Data Science ecosystem and this lack can be a burden to innovation and research in this field.  In this work, we present an open data set of social data content in several Arabic dialects. This data was collected from the Twitter social network and consists on +50K twits in five (5) national dialects. Furthermore, this data was labeled for several applications, namely dialect detection, topic detection and sentiment analysis. We publish this data as an open access data to encourage innovation and encourage other works in the field of NLP for Arabic dialects and social media. A selection of models were built using this data set and are presented in this paper along with their performances.

\end{abstract}
\section{Introduction} 
In last decades, many efforts have been made to enhance Modern Standard Arabic (MSA) Natural Language Processing (NLP). These efforts have led to build systems that can serve more than 400 million people, across Africa and Asia, in many tasks such as machine translation, sentiment classification, diacritization, etc.  However, in most cases, MSA is only used in formal settings, such as newspapers and professional or academic purposes, while Arabic dialects are used in everyday communication.  

In term of resource availability, the majority of Dialectal Arabic (DA) variants are considered as low-resource languages, and suffer from the scarcity of labeled data to build NLP systems. Furthermore, previous works have mainly focused on MSA and some dialects (mainly Egyptian and middle east region dialects) \cite{dahou-etal-2016-word,7422283,mulki-etal-2017-tw}.  Relying on the fact that the MSA and Arabic dialect (DA) variants are etymologically close, the use of MSA NLP systems on DA data has shown shallow performance compared to the performance on MSA data \cite{qwaider-etal-2019-modern}. Thus, in order to enhance the performance of MSA NLP systems and make them generalize better on DA input texts, models should be trained on data containing samples from DA. In other words, there is a real need to open data sets with good quality of labeling for DA. 

Social media (Twitter, Facebook,…) might be the most convenient source to collect DA data, as they provide different contents that reflect the feelings of users across several topics and are written in
the user native and informal Arabic dialect. However, data collected on social media should not be used in its raw form as it suffers from several issues \cite{baly-etal-2017-characterization}. We can cite, for example, the problem of code-switching \cite{elfardy-etal-2014-aida} where users tend to borrow words or phrases from other languages (English or French), which will introduce noise into the collected data especially for the case of automatic annotation. 

The creation of open access social media data sets, such as the one presented in our paper, aims to enhance innovation and practical applications of NLP for DA, such as social media content analysis for marketing studies, public opinion assessment or for social sciences.

In this work, we present the first multi-topic and multi-dialect data set, manually annotated on five (5) DA variants and with five (5) topics. The data set was collected from Twitter and is publicly available and designed to serve several Arabic NLP tasks: sentiment classification, topic classification, and Arabic dialect identification. In order to evaluate the usability of the collected data set, several studies have been performed on it, using different machine learning algorithms such as SVM, Naive Bayes Classifier, etc.  
The main contributions of this paper are:
\begin{itemize}
    \item The introduction of an open source multi-topic multi-dialects corpus for dialectical Arabic.
\item The proof of the usability of our data set through performance evaluation of Arabic dialect identification, Arabic sentiment classification, and Arabic topic categorization systems under different configurations with different machine learning models.

\end{itemize}

As far as we know, the proposed cross-topic and dialect Arabic data set is the first one that covers three tasks at the same time. The rest of the paper is organized as follows. Section \ref{sec:related} discusses some related works. Section \ref{sec:data} presents the collected data and the way it was gathered. The labeling of the data is described on section 4 for the three applications. Finally, section \ref{sec:clc} concludes this paper and gives some outlooks to future work.

%
 
\section{Related works}\label{sec:related}

With the explosion in the number of social media users in the Arab world in recent years, interest in sentiment analysis in Arabic has gained more attention. However, the publicly available data sets are still limited in terms of coverage, size and number of dialects. Moreover, most of the work on Arab sentiment analysis focuses on Modern Standard Arabic, although some authors cover Egyptian and Gulf dialects. On the other hand, performing an analysis of the sentiment of an Arab user on social media relies on many factors (e.g. dialect, topic, …).

\paragraph{Dialectal Arabic sentiment classification.} Recently, several efforts have been made to cover more DA variants. The authors of \cite{9068897} have published an open-access data set which serves to analyze sentiments for Arabic Algerian dialects. The data was mainly 10K comments collected from Facebook pages and manually annotated. The authors applied different experiments using machine learning (e.g. SVM, Naive Bayes) and deep learning methods. Similarly, the authors of \cite{medhaffar-etal-2017-sentiment} have published a publicly available data set of 17K comments collected from Facebook that covers the sentiments and opinions expressed in the Tunisian dialect. Finally, the authors of  \cite{nabil-etal-2015-astd} have proposed a sentiment classification containing 10K comments collected from Twitter on users from multiple Arab countries.

\paragraph{Dialectal Arabic topic classification.} Another direction in research on the analysis of user behaviour is topic classification. \cite{10.5220/0005352102840291} have proposed a publicly available data set containing more than 50K articles from Arabic newspaper websites, distributed on 8 categories (e.g. culture, sport, religion, …). They also proposed some data pre-processing pipelines and an evaluation over several machine learning models (e.g. Decision Tree, Naive Bayes, …). Following the same direction, the authors of \cite{elnagar-etal-2019-automatic}  introduce SANAD, a freely available data set with more than 190K articles that  is used for MSA text categorization. Seven categories where collected from news websites. Despite this interest, no one, to the best of our knowledge, has covered the classification topic on dialectal Arabic.

\paragraph{Arabic dialect identification.} Due to the variety of dialectal Arabic, studying and analysing the behaviour of Arab users and building NLP systems rely heavily on the dialect of the input text. 
As a result, 
considerable work has been conducted recently on the identification of Arabic dialects \cite{bouamor-etal-2019-madar,abdulmageed2020nadi}, and several approaches have been proposed to perform country-level dialect identification and province-level dialect identification. Several models have been proposed in this context: some were based on machine learning models \cite{abu-kwaik-saad-2019-arbdialectid}, while others on advanced deep learning architectures such as BERT pre-trained models \cite{mekki:2020:weighted,talafha2020multidialect}.

\section{Presentation of the data}\label{sec:data}

The data was gathered by randomly scrapping tweets, from active users located in a predefined set of Arab countries, namely : Algeria, Egypt, Lebanon, Tunisia and Morocco. No limits were set for the date of the tweets nor for the exact location in the country. We used Selenium Python library to automate the web navigation and BeautifulSoup to scrap the tweets. The total number of tweets in the data set is 49,306. The tweets distribution per country is given in table \ref{tab:countries}.

\begin{table}[ht]
\begin{center}
\caption{Number of tweets per country}\label{tab:countries}
\begin{tabular}{c|ccccc}
Arabic dialect & Algerian & Lebanon & Morocco & Tunisian & Egyptian \\
\hline
Number of Tweets &13393 & 14482 & 9965 & 8044 & 7519 \\
\hline
\end{tabular}
\end{center}
\end{table}

\section{Data pre-processing Steps}
\subsection{Data cleaning}
In this first step we aimed to remove noise from our data and transform it into a form that is predictable and analyzable for machine learning algorithms. We used mostly regular expressions for the following tasks :
\begin{itemize}
    \item Remove user accounts (@users) to anonymize the data.
    \item Remove Twitter keywords and symbols such as : hastags \#, URLs, RT... from each tweet.
    \item Remove Arabic stop words. We used a list of 250 stop words from NLTK library and +700 words from Github page \cite{stopwordArabic}. We also added few more stop words ourselves.
    \item Remove the emojis. However, we kept the emojis for sentiment analysis because they play a significant role in expressing sentiments.
    \item Remove punctuation characters. We used the list from the NLTK library and we included Arabic punctuation characters, such as the reverse question mark in Arabic.
    \item Remove words not written in Arabic for dialect detection. e.g. “Thank you”.
    \item Calculate the length of Tweets and remove meaningless short Tweets.
\end{itemize}

\subsection{Stemming/Lemmatization}

Stemming is the process of reducing a word to its word stem that affixes to suffixes and prefixes. For example: changing, by removing suffix “ing” their stem will be “chang'' without “e”.

Lemmatization consists of representing words in their root form (lemma). For a verb, it will be its infinitive. For a name, its singular masculine form. The idea is once again to keep only the meaning of the words used in the corpus. For example: “gone” and “went“ their root is “go”.

We used stemming for words with unknown root in Arabic words (Like \AR{برشا} in Tunisian dialect).

\subsection{Vectorisation}

Machine Learning algorithms are applied to numeric data. Hence, to transform text data to numbers, we used TFIDF (Term Frequency–Inverse Document Frequency) and Bag-of-Words (BOW) techniques. This last is the simplest representation that turns arbitrary text into fixed-length vectors by counting how many times each word appears. The common BOW vectorizer used is Countvectorizer from Scikit-learn Python package.

\begin{table}
\caption{Example of TFIDF vectorisation} \label{tab:}
\begin{center}
\begin{tabular}{c|cccccccccc}
& for &great & greatest & lasagna & life & love & loved & the & thing & times   \\
\hline
sentence1 & 0 &0 & 1 & 0 & 1 &1 &0& 1 & 1 & 0\\
\hline
sentence2 & 0 &2 & 0 & 0 & 0 &1 &1& 0 & 0 & 0\\
\hline
sentence3 & 0 &0 & 1 & 0 & 0 &1 &0& 1 & 1 & 0\\
\hline
sentence4 & 1 &0 & 0 & 1 & 0 &1 &0& 0 & 0 & 1\\
\hline
\end{tabular}
\end{center}
\end{table}

TFIDF converts a collection of raw documents to a matrix of TF-IDF features. It reflects how important a word is to a document in a collection, in our example a collection of Tweets. The scores are computed as follows :
$$ W_{x,y}=\text{tf}_{x,y} \times log\left(\frac{N}{\text{df}_x}\right)$$
with $\text{tf}_{x,y}$ the frequency  of word $x$ in document $y$, $\text{df}_{x}$ the number of documents containing word $x$ and $N$ the total number of documents.

\section{Data Labeling}
\subsection{Semi-automatic Data Labeling process}

In order to label the tweets, we first used the MonkeyLearn tool to manually label one thousand (1000) Tweets for each application (Dialect, Topic and Sentiment detection). We then built multiple models based on these labeled tweets. In the next step, we predicted the labels for 200 non-labeled tweets using the built models, and then corrected manually the ones that were labeled incorrectly. We rebuilt the models by adding the new tweets (1200 tweets in the second time) and repeated the process by increasing the number of non-labeled tweets to predict until we finish and check the labeling for all the data set.

In the following subsections we present the main statistics for the data after labeling.

\subsection{Topic detection}

\begin{figure}[h!]
\centering
\includegraphics[scale=0.37]{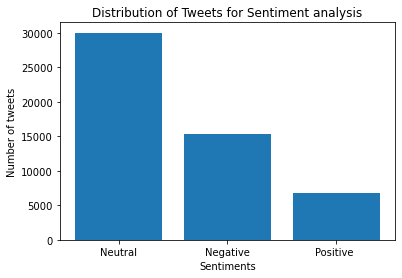} \includegraphics[scale=0.37]{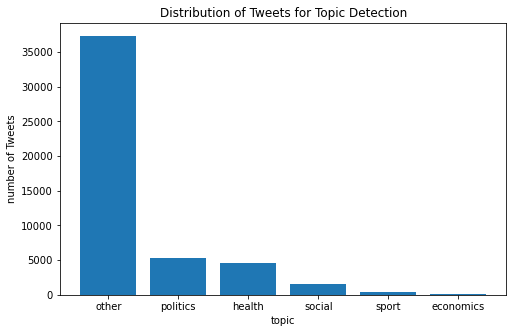}
\caption{Distribution of Tweets for Sentiment analysis (left) and Topic detection (right). }
\label{fig:topics}
\end{figure}

We choose to attach topic labels to tweets among one of the following topics : Politics, Health, Social, Sport, Economics. The tweets that were not relevant to one of these labels were labeled as 'Other' as shown in table \ref{tab:topics}. Only $25\%$ of the tweets were labeled to one of the topic categories and $75\%$ were labeled such as 'Other' topic. However, we believe that keeping the 'Other' label category is important for topic detection application, so we publish the labeled data set as such. Given the unbalanced character of the labeled data (see fig \ref{fig:topics}), we recommend the use of under-sampling (i.e. using only a part of the data) for the 'Other' topic category when building machine leaning models. 

\begin{table} \caption{Number of tweets per topic}
\label{tab:topics}
\begin{center}
\begin{tabular}{c|cccccc} 
Topic & Other & Politics & Health & Social & Sport & Economics \\
\hline
Number of Tweets &37313 & 5355 & 4574 & 1564 & 94 & 406 \\
\hline
\end{tabular}
\end{center}
\end{table}


\subsection{Dialect detection}

\begin{figure}[h!]
\centering
\includegraphics[scale=0.5]{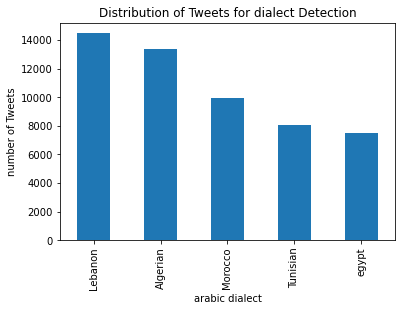}
\caption{Distribution of Tweets for dialect Detection.}
\label{hist:countries}
\end{figure}

The number of tweets per dialect is given in table \ref{tab:countries}. In figure \ref{hist:countries} we present the same data in a histogram. The number of tweet per dialect ranges between eight thousands (8000) and fourteen thousands (14000). 

\subsection{Sentiment analysis:}

The total number of tweets that we labeled for sentiment analysis is 52,210 tweets. The majority of tweets were labeled with a neutral sentiment.

\begin{table}[h!]
\caption{Number of tweets per label for sentiment analysis}
\begin{center}
\begin{tabular}{cccc}
Sentiment Analysis& Positive & Negative & Neutral  \\
\hline
Number of Tweets &6792 & 15385 & 30033 \\
\hline
\end{tabular}
\end{center}
\end{table}

\section{Baseline models}\label{sec:models}

Using our labeled data, we evaluate the performance of Arabic dialect identification, Arabic sentiment classification, and Arabic topic categorization systems with the following  machine learning models : Logistic Regression, SGD Classifier, Linear SVC and Naive bayes. We used grid search and pipelines to find the best hyper-parametres.

\subsection{Dialect detection}

As can be seen in table \ref{tab:results_dialects}, the Naive bayes algorithm performs better than the other models on our testing set.
\begin{table}
\caption{Performance of the different tested models for dialect detection}
\label{tab:results_dialects}
\begin{center}
\begin{tabular}{c|cccc}
models &SGD Classifier & Logistic Regression & Naive bayes & Linear SVC \\
\hline
f1-score(micro) &0.73 &0.72 &0.75 &0.75 \\
\hline
precision &0.75 &0.72 & 0.80 &0.76 \\
\hline
recall &0.72 &0.71 &0.71 &0.74 \\
\hline
accuracy &0.72 &0.72 & 0.75 &0.76 \\
\hline
balanced accuracy &0.71 &0.71 & 0.79 &0.75 \\
\hline
\end{tabular}
\end{center}
\end{table}

\subsection{Topic detection}

As shown in table \ref{tab:results_topic}, Logistic Regression and SGD Classifier both have a good performance, in our case (imbalanced Data) we chose Logistic Regression because of its high f1-score value. We recall that for topic detection we recommend to use under-sampling of the 'other' category as it is over-represented in our data set. 

\begin{table}
\caption{Performance of the different tested models for topic detection}
\label{tab:results_topic}
\begin{center}
\begin{tabular}{ccccc}
models &Logistic Regression &SGD Classifier & Linear SVC \\
\hline
f1-score(micro) &0.82 &0.72 &0.84 \\
\hline
precision &0.59 &0.65 & 0.65 \\
\hline
recall &0.59 &0.45 &0.45 \\
\hline
accuracy &0.82 &0.84 & 0.84 \\
\hline
balanced accuracy &0.70 &0.78 & 0.78 \\
\hline
\end{tabular}
\end{center}
\end{table}

\subsection{Sentiment analysis}

According to the results in table \ref{tab:results_sentiments}, the four models built have similar results for sentiment analysis, except Naive bayes which is less efficient. The SGD Classifier out-performes slightly in term of f1-score and accuracy. 

\begin{table}
\caption{Performance of the different tested models for Sentiment Analysis}
\label{tab:results_sentiments}
\begin{center}
\begin{tabular}{ccccc}
models & SGD Classifier & Logistic Regression & Naive bayes & Linear SVC \\
\hline
f1-score(micro) &0.74 &0.73 &0.67 &0.73 \\
\hline
precision &0.75 &0.75 &0.75 &0.75 \\
\hline
recall &0.76 &0.74 &0.64 &0.75 \\
\hline
accuracy &0.77 &0.76 &0.73 &0.76 \\
\hline
balanced accuracy &0.74 &0.74 &0.63 &0.74 \\
\hline
\end{tabular}
\end{center}
\end{table}

\section{Conclusion}\label{sec:clc}

In this work we presented a Labeled data set of ~50K tweets in five (5) Arabic dialects. We presented the process of labeling this data for dialect detection, topic detection and sentiment analysis. We put this labeled data openly available for the research, startup and industrial community to build models for applications related to NLP for dialectal Arabic. We believe that initiatives such as ours can catalyse the innovation and the technological development of AI solutions in Arab countries like Morocco by removing the burden linked to the non availability of labeled data and to time-consuming tasks of collecting and manually labeling data. 
We also presented a set of Machine learning models that can be used as baseline models and to which future users of this data set can compare and aim to outperform by innovating in term of computational methods and algorithms. The labeled data set can be downloaded at  \cite{msda_datasets}, and all the implemented algorithms are available at \cite{codeGithub}.

%
%
%

\bibliographystyle{splncs04}
\bibliography{mybibliography}

\end{document}